\documentclass[11pt,a4paper]{article}

\usepackage[margin=2.5cm]{geometry}
\usepackage{graphicx}
\usepackage{booktabs}
\usepackage{multirow}
\usepackage{array}
\usepackage{longtable}
\usepackage{tikz}
\usepackage{tikzdotncross}
\usetikzlibrary{positioning,calc,arrows.meta}

\usepackage{amsmath}
\usepackage{amssymb}

\usepackage{enumitem}

\usepackage{hyperref}
\usepackage[nameinlink]{cleveref}

\usepackage[numbers]{natbib}

\begin{document}
\title{Academic League of Artificial Intelligence --  \\ 
An Integrative Perspective of Teaching, Research, and Extension}
%
\date{}
\author{
Alison R. Panisson* \and Maria Eduarda W. M. Vianna \and Italo Firmino da Silva \and Heitor Henrique da Silva \and Rafaela Fernandes Savaris \and Bernardo Pandolfi Costa \and Martin Augusto Gagliotti Vigil \and Jim Lau \and Agenor Hentz \and Andréa Sabedra Bordin \and Alexandre Leopoldo Gonçalves \and Roberto Rodrigues-Filho \\ \\
Federal University of Santa Catarina (UFSC), Brazil \\
\texttt{* alison.panisson@ufsc.br}
}

\maketitle             
\begin{abstract}

Academic leagues have become important mechanisms for promoting extracurricular education and strengthening the integration between universities and society. This paper presents the organizational framework adopted by the Academic League of Artificial Intelligence (LIA) at the Federal University of Santa Catarina (UFSC), designed to integrate teaching, research, and university extension through a student-centered, project-based approach. The framework combines democratic governance, collaborative learning, and dynamic project organization to foster both technical and transversal competencies. 
The framework is illustrated through representative initiatives, including competition teams, study groups, open lectures, knowledge repositories, and AI-powered applications with social impact. These projects demonstrate how diverse educational, scientific, and extension activities can be developed within a common organizational structure while promoting leadership, scientific production, community engagement, and knowledge preservation.
The reported experience indicates that the proposed framework provides a flexible and replicable model for integrating the three university pillars into engineering and computing education, offering practical guidance for academic leagues and similar student organizations.

\end{abstract}

\section{Introduction}

Since the 1988 Brazilian Constitution, teaching, research, and extension have been established as inseparable pillars of higher education in Brazil. This principle states that university education should not be fragmented but instead conceived as a holistic process in which theoretical learning (teaching), knowledge creation (research), and social engagement (extension) are continuously articulated~\cite{gimenez2024three}.

Among these pillars, university extension plays an important role by connecting academic knowledge with societal demands. Extension activities enable students to apply theoretical concepts and research outcomes in real-world contexts while fostering leadership, communication, teamwork, and social responsibility~\cite{oliveira2024transformative}. Rather than representing an isolated university mission, extension serves as the bridge between teaching and research, allowing knowledge produced within universities to generate concrete impacts on society.

Research also plays a fundamental role in student education by developing critical thinking, scientific reasoning, and investigative skills. Undergraduate research projects often constitute students' first contact with scientific inquiry, providing experiences that subsequent enrich both classroom learning and extension activities~\cite{Guimaraes2024}. In turn, extension activities create authentic scenarios in which research questions emerge from interactions with society, reinforcing the complementary relationship among the three university pillars.

This multidimensional perspective of higher education has motivated Brazilian universities to create extracurricular initiatives capable of integrating teaching, research, and extension into coherent educational experiences~\cite{CNECES2018Resolucao7}. Among these initiatives, academic leagues have become particularly successful. Originally popularized in health-related fields, academic leagues are student-led organizations supervised by faculty members that combine theoretical education, research activities, and community engagement in long-term projects. These organizations promote student autonomy while contributing to professional development and the university's social mission~\cite{de2024context}.

Although academic leagues have been extensively studied in medicine and other health sciences, there is still very limited literature describing their adoption in Computer Science and Engineering. Likewise, little attention has been given to how these organizations can be structured to integrate the three university pillars while promoting student protagonism, technical excellence, and community engagement. As Artificial Intelligence rapidly becomes one of the most strategic areas of computing, understanding how academic leagues can contribute to AI education and outreach becomes increasingly relevant.

This paper addresses this gap by presenting the organizational model adopted by the Academic League of Artificial Intelligence (LIA) at the Federal University of Santa Catarina (UFSC). Rather than simply reporting isolated extension activities, we describe a student-centered framework that organizes projects around the continuous integration of teaching, research, and extension\footnote{Fulfilling one of the fundamental principles of university extension, i.e., the inseparability of teaching, research, and extension, as established by the Brazilian National Extension Plan~\cite{forproex2012}.}. 
Although student organizations in computing frequently involve undergraduate students, they are typically organized around specific research agendas or technological developments rather than around an educational framework explicitly designed to integrate the three university pillars. Consequently, documented organizational models that can guide the creation, management, and evaluation of academic leagues in Computer Science and Artificial Intelligence remain scarce. The proposed framework is illustrated through representative initiatives developed within LIA, including competition teams, study groups, open lecture series, knowledge repositories, and community outreach activities. By documenting this organizational experience, the paper provides a reference model that can support the establishment of similar initiatives at other institutions while contributing to the emerging discussion on student-centered educational structures in computing.

The main contributions of this paper are threefold: (i) we discuss the role of academic leagues as organizational structures for integrating teaching, research, and extension within Computer Science and Artificial Intelligence education. To the best of our knowledge, this is the first paper to examine academic leagues as an organizational framework for student-centered education in these fields; (ii) we present the organizational framework adopted by LIA, emphasizing student protagonism, project organization, and the articulation of the three university pillars; and (iii) we report representative projects developed within the league, discussing the educational outcomes, extension products, and lessons learned throughout its implementation.

The remainder of this paper is organized as follows. Section~\ref{sec:background} reviews the role of academic leagues in Brazilian higher education and identifies the existing gap in Computing and Engineering. Section~\ref{sec;lia} presents the proposed organizational framework and its implementation within the Academic League of Artificial Intelligence (LIA). Section~\ref{sec:projects} describes representative projects developed within the league to illustrate the framework in practice. Section~\ref{sec:lessons} discusses the organizational lessons learned from its implementation. Finally, the paper concludes by summarizing the main contributions, discussing the framework's applicability to other contexts, and outlining directions for future work.

\section{Background}\label{sec:background}

The Brazilian higher education system is founded on the principle that teaching, research, and extension should be developed in an integrated and inseparable manner~\cite{brasil1988constituicao}. Consequently, educational initiatives capable of simultaneously articulating these three dimensions have become increasingly important for promoting comprehensive student development.

Among the different institutional mechanisms adopted by Brazilian universities, we believe that academic leagues have emerged as successful student-centered organizational models. Although originally established and studied in health-related fields, academic leagues have also expanded to other areas of knowledge. Their organizational structure encourages students to participate in activities that combine formal education, scientific research, and community engagement under faculty supervision. 
This section reviews the role of academic leagues within the Brazilian higher education system and discusses their relationship with the three university pillars, highlighting the existing gap in computing and artificial intelligence.

\subsection{Academic Leagues in Brazil}

The literature consistently characterizes academic leagues as student-led organizations that integrate teaching, research, and extension through continuous educational and community-oriented activities. Most published studies focus on medical and health-related leagues (e.g.,~\cite{Camilo2020,DosSantosSilva2021}), where these organizations are recognized as important instruments for professional education and university extension.

Academic leagues are student-led organizations composed of undergraduate students who share a common interest in a particular topic. According to the systematic review by Fernandes et al.~\cite{Fernandes2025}, under faculty supervision they develop teaching, research, and extension activities that strengthen the integration between universities and society.

Furthermore, Goergen et al.~\cite{goergen2023exploratory} characterize academic leagues as extracurricular initiatives that combine theoretical instruction, practical activities, scientific research, and outreach projects under faculty supervision. Similarly, Matheus et al.~\cite{matheus2019important} describe them as organizations dedicated to promoting education, research, and community assistance, while tracing their origins in Brazil to the \textit{Liga de Combate à Sífilis} (Academic League Against Syphilis), founded in 1920.

Other studies reinforce this perspective by describing academic leagues as long-term university extension programs that foster student autonomy, multidisciplinary collaboration, and professional development while maintaining a strong connection with society~\cite{santos2020strengthening,cavalcante2021search}. 
Further,~\cite{oliveira2024transformative} highlights student participation in extension projects as a key mechanism for integrating teaching, research, and community engagement, an educational philosophy that closely aligns with the organizational model adopted by academic leagues.

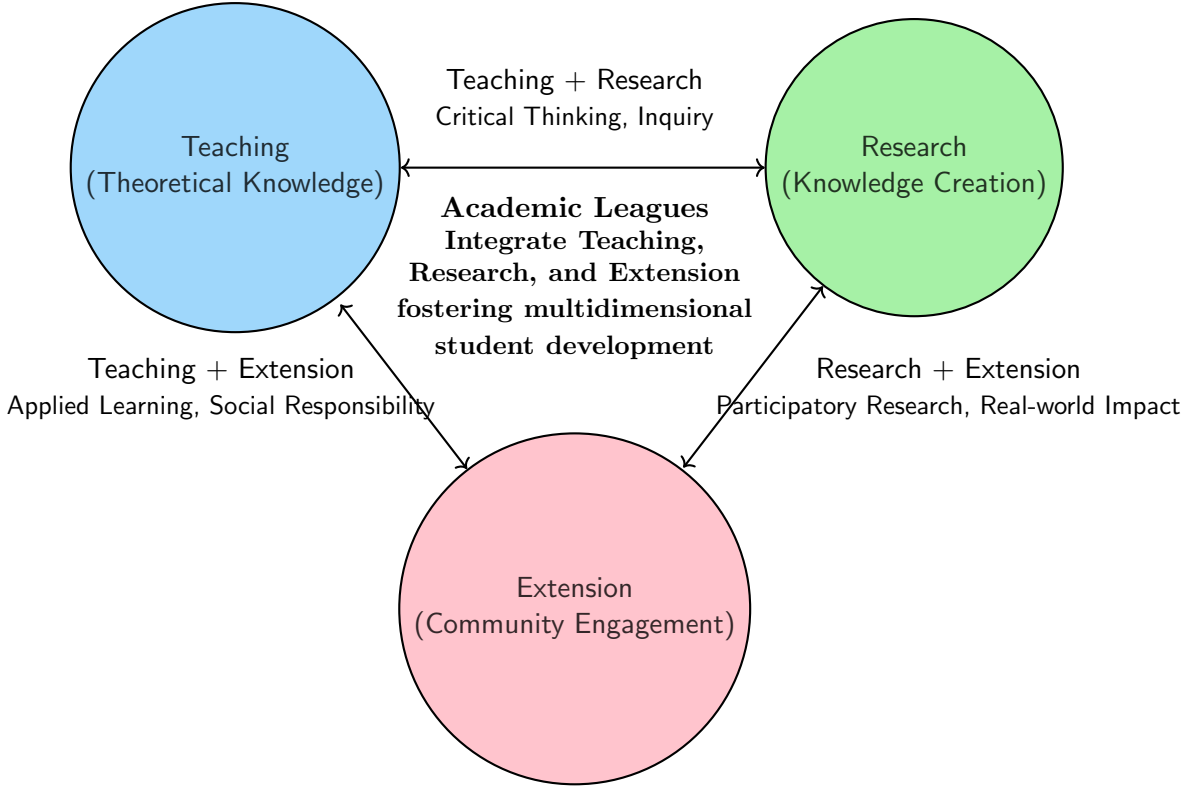
\begin{figure}[t]
\begin{tikzpicture}[thick, every node/.style={font=\sffamily}, scale=0.9]

\definecolor{teachingcolor}{RGB}{135,206,250}   
\definecolor{researchcolor}{RGB}{144,238,144}   
\definecolor{extensioncolor}{RGB}{255,182,193}  

\node[draw, circle, fill=teachingcolor, fill opacity=0.8, minimum size=3cm, align=center] (teaching) {Teaching \\ (Theoretical Knowledge)};
\node[draw, circle, fill=researchcolor, fill opacity=0.8, minimum size=3cm, align=center, right=7cm of teaching.center] (research) {Research \\ (Knowledge Creation)};
\node[draw, circle, fill=extensioncolor, fill opacity=0.8, minimum size=3cm, align=center, below=3.5cm of $(teaching)!0.5!(research)$] (extension) {Extension \\ (Community Engagement)};

\node[align=center] at ($(teaching)!0.5!(research) + (0,1.0)$) {Teaching + Research\\ \small{Critical Thinking, Inquiry}};
\node[align=center] at ($(teaching)!0.5!(extension) + (-2.7,0)$) {Teaching + Extension\\ \small{Applied Learning, Social Responsibility}};
\node[align=center] at ($(research)!0.5!(extension) + (3,0)$) {Research + Extension\\ \small{Participatory Research, Real-world Impact}};

\node[align=center, font=\bfseries, text width=5cm] at ($(teaching)!0.5!(research)!0.25!(extension)$) {Academic Leagues \\ \small{Integrate Teaching, Research, and Extension \\ fostering multidimensional student development}};

\draw[<->, thick] (teaching) -- ($(teaching)!0.78!(research)$);
\draw[<->, thick] (research) -- ($(research)!0.68!(extension)$);
\draw[<->, thick] (extension) -- ($(extension)!0.69!(teaching)$);

\end{tikzpicture}
\caption{Academic Leagues as integrative organization of teaching, research, and extension.}
\label{fig:pilars}
\end{figure}

Overall, the literature characterizes academic leagues as institutional mechanisms that effectively integrate teaching, research, and extension through student-centered activities. By promoting the continuous interaction among these three university pillars, academic leagues foster active learning, student protagonism, and meaningful engagement with society, ultimately contributing to more comprehensive educational outcomes~\cite{souza2022third}. Figure~\ref{fig:pilars} 
provides an overview of the positioning of Academic Leagues as integrative organization of teaching, research, and extension.

\subsection{Academic Leagues in Computer Science and Engineering}

In contrast to the extensive literature on academic leagues in health sciences, there is very limited academic evidence regarding their adoption in Computer Science and Engineering. Our review did not identify published studies describing the organization, operation, or educational impact of academic leagues in these areas.

Student organizations are common in computing programs, including ACM Student Chapters, IEEE Student Branches, programming competition teams, robotics groups, and hackathon clubs. These initiatives contribute significantly to students' technical and professional development by promoting extracurricular learning, teamwork, and interaction with industry. However, they are rarely characterized or studied as institutional mechanisms for integrating teaching, research, and university extension.

Similarly, several universities maintain research laboratories and extension projects in Artificial Intelligence, software engineering, robotics, and related areas. Although these initiatives frequently involve undergraduate students, they are generally organized around specific research agendas or technological developments rather than around a student-centered educational framework integrating the three university pillars.

This lack of documented organizational models makes it difficult for universities to establish new academic leagues in computing or to evaluate their educational contributions. While the concept of academic leagues is well consolidated in health sciences, its adaptation to Computer Science and Artificial Intelligence remains unexplored in the literature.

Table~\ref{tab:academic_leagues} presents a mapping of academic organizations related to Artificial Intelligence, Computing, and associated fields in Brazil as of 2025. To the best of our knowledge, no comprehensive mapping of these organizations has been reported in the scientific literature. Therefore, the mapping was compiled from publicly available sources, including institutional websites, social media pages, and other online resources.


\begin{table}[htbp]
\centering
\small
\setlength{\tabcolsep}{3pt}
\renewcommand{\arraystretch}{1.05}
\resizebox{\columnwidth}{!}{%
\begin{tabular}{p{3.8cm}|p{2.3cm}|p{1.2cm}|p{5.0cm}}
\hline
\textbf{League Name} & \textbf{University} & \textbf{Year} & \textbf{Main Focus / Activities} \\
\hline \hline
LIA -- Liga Acadêmica de Inteligência Artificial
& UFSC--Araranguá & 2021
& Artificial intelligence, research, extension activities, workshops, and student projects \\
\hline
LADS -- League of Artificial Intelligence and Data Science
& UNAMA & 2025
& Artificial intelligence and data science, with emphasis on student development and practical activities \\
\hline
TAIL -- Technology and Artificial Intelligence League
& UFPB & 2020
& Artificial intelligence, technology development, student projects, and academic activities \\
\hline
DATA -- Inteligência Artificial e Ciência de Dados
& USP--São Carlos & 2019
& Machine learning, artificial intelligence, data science, and related technical activities \\
\hline
LIIA -- Liga de Inovação e Inteligência Artificial
& UFSM--Santa Maria & --
& Artificial intelligence, innovation, interdisciplinary activities, and technology development \\
\hline
LIGIA -- Liga Acadêmica de Inteligência Artificial
& UFPE & --
& Artificial intelligence and deep learning, with academic and practical activities \\
\hline
LACDA -- Liga Acadêmica de Ciência de Dados Aplicada
& USP--São Paulo & 2021
& Applied data science, data analysis, machine learning, and practical projects \\
\hline
LAIA -- Liga Acadêmica de Inteligência Artificial
& UFS--Sergipe & --
& Artificial intelligence, algorithms, computational methods, and student activities \\
\hline
LIAO -- Liga Acadêmica de IA e Otimização
& UFBA--Salvador & --
& Artificial intelligence, machine learning, optimization, and computational problem solving \\
\hline
LINCE -- Liga de Inteligência Neuro-computacional na Engenharia
& UNESP--Sorocaba & 2023
& Artificial intelligence, neurocomputing, computational intelligence, and engineering applications \\
\hline \hline
\end{tabular}%
}
\caption{Academic leagues focused on artificial intelligence, data science, and related computational fields in Brazilian universities.}
\label{tab:academic_leagues}
\end{table}

\section{Organizational Framework for Academic Leagues}\label{sec;lia}

This paper proposes a student-centered organizational framework for academic leagues in the fields of Computer Science and Engineering. Rather than defining a fixed set of projects, the framework establishes an organizational structure in which different initiatives are developed as complementary mechanisms for integrating the three university pillars. The framework has been adopted and refined through the Academic League of Artificial Intelligence (LIA) at the Federal University of Santa Catarina (UFSC), serving as the basis for all projects developed within the league.

The central premise of the proposed framework is that students should occupy a leading role in both the conception and execution of projects, while faculty members act primarily as mentors and facilitators. 
Consequently, every project developed within the league is expected to simultaneously promote student learning, generate technical or scientific knowledge, and produce extension outcomes directed toward the broader community.

\subsection{Institutional Context}

The Academic League of Artificial Intelligence (LIA) was established at the Araranguá Campus of the UFSC with the objective of promoting Artificial Intelligence education through the continuous integration of teaching, research, and university extension. 
Unlike research laboratories dedicated to specific scientific topics or student organizations focused on isolated extracurricular activities, LIA was conceived as an institutional environment in which students could progressively develop both technical and transversal competencies while actively participating in projects with educational and social impact.

Since its creation in 2019, the league has developed initiatives in areas including Artificial Intelligence education, data analysis and visualization, machine learning, computer vision, multi-agent systems, programming competitions, open educational resources, offensive security, and community outreach. 
These projects differ in their technical objectives but share the same organizational principles, allowing students to move between teaching, research, and extension activities throughout their academic journey.

\subsection{Organizational Structure of LIA}

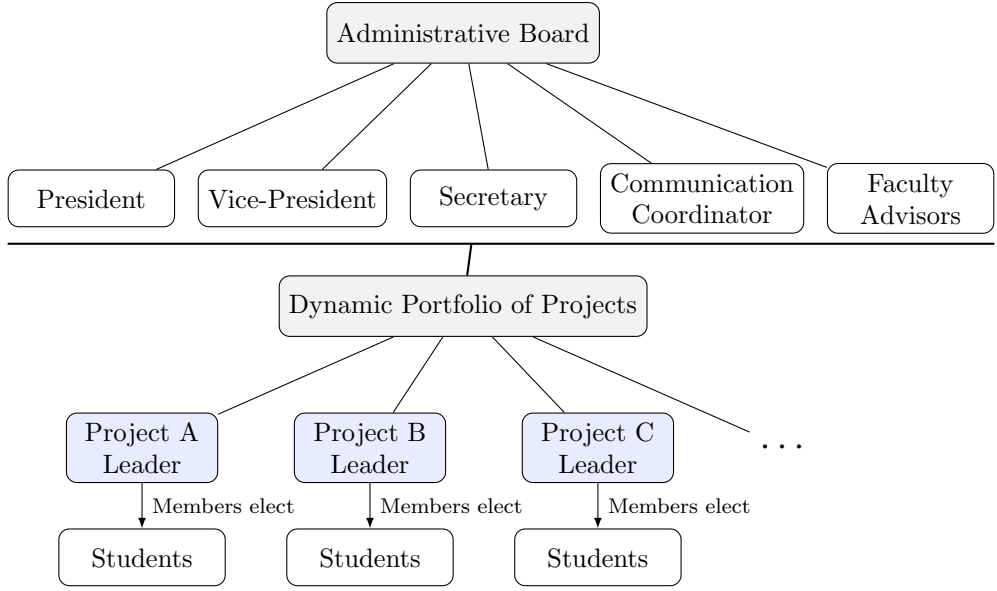
\begin{figure}[t]
\centering
\begin{tikzpicture}[
    >=latex,
    node distance=1.0cm,
    every node/.style={font=\small},
    box/.style={
        draw,
        rounded corners,
        minimum width=2.8cm,
        minimum height=0.8cm,
        align=center,
        fill=gray!10
    },
    role/.style={
        draw,
        rounded corners,
        minimum width=2.2cm,
        minimum height=0.75cm,
        align=center
    },
    project/.style={
        draw,
        rounded corners,
        minimum width=2.0cm,
        minimum height=0.75cm,
        align=center,
        fill=blue!8
    }
]

\centering
\node[box] (board)
{Administrative Board};

\node[role, below left=1.4cm and 2cm of board] (pres)
{President};
\node[role, right=0.3cm of pres] (vice)
{Vice-President};
\node[role, right=0.3cm of vice] (sec)
{Secretary};
\node[role, right=0.3cm of sec] (comm)
{Communication\\Coordinator};
\node[role, right=0.3cm of comm] (adv)
{Faculty\\Advisors};
\foreach \x in {pres,vice,sec,comm,adv}
    \draw[-] (board) -- (\x);

\coordinate[below=0.6cm of pres.west] (barL);
\coordinate[below=0.6cm of adv.east] (barR);

\draw[thick] (barL) -- (barR);

\node[box, below=2.8cm of board] (projects)
{Dynamic Portfolio of Projects};
\coordinate (mid) at ($(barL)!0.47!(barR)$);
\draw[thick] (mid) -- (projects);

\node[project, below left=1cm and 0.8cm of projects] (p1)
{Project A\\Leader};
\node[project, right=1cm of p1] (p2)
{Project B\\Leader};
\node[project, right=1cm of p2] (p3)
{Project C\\Leader};
\node[right=1cm of p3] (dots) {\Large$\cdots$};

\foreach \x in {p1,p2,p3,dots}
    \draw[-] (projects) -- (\x);

\node[role, below=0.6cm of p1] (s1)
{Students};
\node[role, below=0.6cm of p2] (s2)
{Students};
\node[role, below=0.6cm of p3] (s3)
{Students};
\foreach \x/\y in {p1/s1,p2/s2,p3/s3}
    \draw[->] (\x) -- node[right,font=\scriptsize]{Members elect} (\y);

\end{tikzpicture}
\caption{Organizational structure adopted by LIA.}
\label{fig:lia_structure}
\end{figure}

The proposed organizational framework combines a stable administrative structure with a dynamic portfolio of projects as illustrated in Figure~\ref{fig:lia_structure}. The administrative structure is responsible for ensuring the continuity of the academic league across successive generations of students, while projects constitute the operational units through which teaching, research, and extension activities are carried out. 

The administrative board is composed of students and is organized into predefined roles, including a President, Vice-President, Secretary, and Communication Coordinator. These positions are responsible for the overall management of the league, including strategic planning, member recruitment, institutional communication, event organization, documentation, and interactions with the university administration. Faculty members act primarily as advisors, providing institutional and academic guidance while allowing students to assume the leading role in the management of the league.

The administrative board is elected annually by the members of the league through a democratic process in which groups of candidates present their management proposals and organizational plans. To ensure continuity in the league's administration, only students expected to remain enrolled at the university for at least one additional year are eligible to run for these positions. This criterion allows elected members to complete their one-year term while facilitating the planning and execution of long-term initiatives. The election process encourages members to actively participate in defining the strategic direction of the league while fostering leadership, responsibility, transparency, and collaborative decision-making.

Unlike the administrative board, projects have a flexible and dynamic organization. New projects may be created according to emerging opportunities, student interests, research demands, funding opportunities, or extension initiatives, while completed projects may be concluded or incorporated into other activities. Each project is coordinated by a student project leader, who is responsible for planning activities, organizing the project team, monitoring progress, and reporting results to the administrative board.

Project leaders are elected by the members participating in each project. Although every member is eligible to assume this role, leadership naturally tends to be entrusted to the most experienced students, who have accumulated technical expertise and organizational experience through previous participation in the league. This election process promotes the gradual transfer of knowledge between generations of students while recognizing leadership as a responsibility earned through experience and engagement rather than assigned by faculty members.

This separation between administrative management and project execution has proven essential for the sustainability of the league. While the administrative board preserves the institutional identity, organizational processes, and long-term planning of LIA, projects remain flexible to evolve according to technological advances, student interests, and community demands. 
To ensure effective coordination, weekly meetings are held with the leaders of all active projects. These meetings provide a forum for reporting progress, discussing challenges, exchanging experiences, coordinating joint activities, identifying opportunities for collaboration, recruiting volunteers to support league-wide initiatives, and planning future actions. This continuous communication mechanism strengthens collaboration among projects while preserving their autonomy. 
Consequently, the league is able to continuously renew its portfolio of activities while maintaining a stable and well-coordinated governance model.

\subsection{Project Development Process}

\begin{figure}[htb]
\centering
\begin{tikzpicture}[
    node distance=2.2cm,
    every node/.style={font=\small},
    box/.style={
        draw,
        rounded corners,
        align=center,
        minimum width=1cm,
        minimum height=0.7cm,
        fill=blue!5
    },
    circlebox/.style={
        draw,
        circle,
        minimum size=2cm,
        align=center,
        fill=green!10
    },
    >={Latex[length=2mm]},
    arrow/.style={thick, ->}
]
\definecolor{teachingcolor}{RGB}{135,206,250}   
\definecolor{researchcolor}{RGB}{144,238,144}   
\definecolor{extensioncolor}{RGB}{255,182,193}  
\node[box] (sources)
{
New projects originate from previous activities\\ or external opportunities
};
\node[box, below=0.5cm of sources] (approval)
{
Administrative board approves new projects
};
\node[box, below=0.5cm of approval] (leader)
{
Participants are called and a leader is elected
};
\node[box, below=0.5cm of leader] (study)
{
Study groups investigate topics
};
\node[circlebox, fill=researchcolor, fill opacity=1, below=.75cm of study] (research)
{Research};
\node[circlebox, fill=teachingcolor, fill opacity=1, left=1cm of research] (teaching)
{Teaching};
\node[circlebox, fill=extensioncolor, fill opacity=1, right=1cm of research] (extension)
{Extension};
\node[box, below=.75cm of research] (products)
{
Knowledge dissemination
};
\draw[arrow] (sources) -- (approval);
\draw[arrow] (approval) -- (leader);
\draw[arrow] (leader) -- (study);
\draw[arrow] (study) -- (research);

\draw[<->] (teaching) to[bend left=0] (research);
\draw[<->,] (research) to[bend left=0] (extension);
\draw[<->, name path=teachext] (extension) to[bend left=35] (teaching);

\path[name path=vertline] (research) -- (products);
\path[name intersections={of=vertline and teachext, by=hop}];

\draw[arrow] (research) -- ($(hop)+(0,0.1)$)
    arc[start angle=90, end angle=-90, radius=0.1]
    -- (products);

\draw[arrow] (study) -- (teaching);
\draw[arrow] (study) -- (extension);

\draw[arrow] (teaching.south) -- (products.north west);
\draw[arrow] (extension.south) -- (products.north east);

\draw[arrow]
(products.east) -|
++(3,0)
|- (sources.east);

\node[box, below=.75cm of products] (documentation)
{
Systematic documentation
};

\draw[arrow] (products) -- (documentation);

\draw[arrow]
(documentation.west) -|
++(-2.9,0)
|- (study.west);

\end{tikzpicture}
\caption{An overview of the project development process.}
\label{fig:projectdevprocess}
\end{figure}
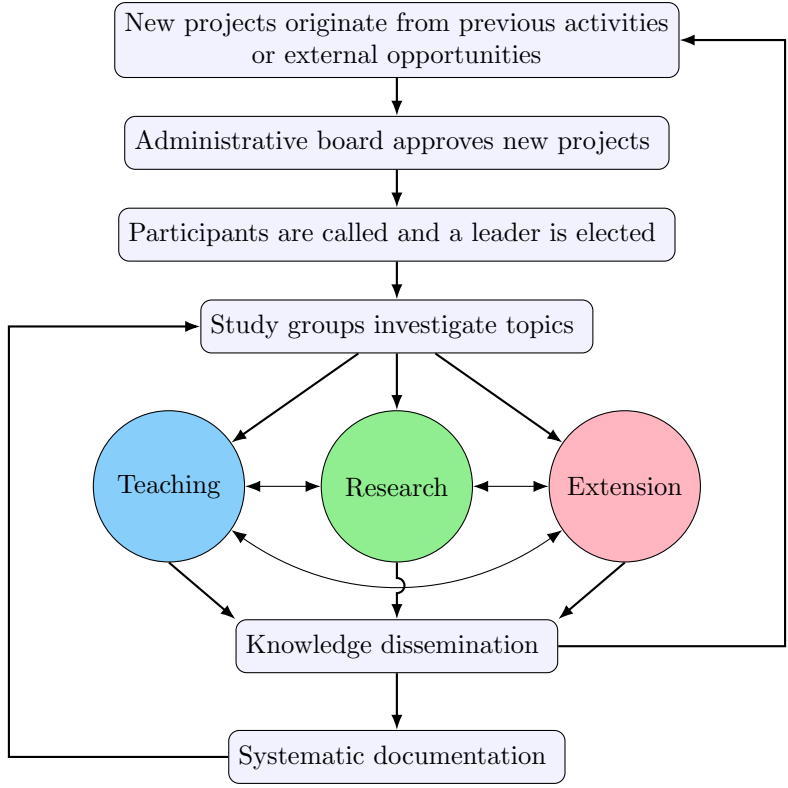

The organizational framework adopted by LIA is centered on projects. Rather than defining a fixed portfolio of activities, the league continuously creates, evolves, and concludes projects according to students' interests, emerging research opportunities, community demands, and institutional partnerships. This dynamic organization enables the league to adapt to emerging technological trends, in particular the fast-evolving field of Artificial Intelligence, while maintaining a stable governance structure. The steps of such a process are illustrated in Figure~\ref{fig:projectdevprocess} and detailed in the following.

Projects may originate from different sources, including undergraduate and graduate courses, scientific initiation projects, research collaborations, extension proposals, student initiatives, or external opportunities such as competitions and funding calls. Regardless of their origin, all projects are expected to contribute, to different extents, to the integration of teaching, research, and extension.

Once a project is approved by the administrative board, the administrative board creates a call for participants of the project, then participating students elect a project leader responsible for coordinating the activities and serving as the primary point of contact between the project team and the administrative board. The project leader is responsible for organizing meetings, planning activities, monitoring progress, and coordinating the production of project deliverables.

During the early stages of project development, students are encouraged to organize study groups to investigate the scientific literature, existing technological solutions, and state-of-the-art approaches related to the project. 
This phase establishes a common technical foundation among participants while promoting collaborative learning and critical discussion. 
To support this process, study groups are encouraged to leverage the educational resources available in LIA's repositories, including tutorials, technical documentation, and implementation examples developed in previous projects. 
This practice facilitates the onboarding of new members, promotes knowledge reuse, and enables students to build upon the experience accumulated by earlier generations of participants.
In fact, the continuous transformation of learning activities into research outputs and extension products constitutes one of the defining characteristics of the proposed organizational framework. Rather than treating teaching, research, and extension as independent activities, projects naturally evolve through these three dimensions, creating a continuous cycle of knowledge generation, application, and dissemination.

As projects mature, their initiatives are transformed into educational, scientific, and extension outputs. Depending on the nature of the project, these products may include technical reports, scientific publications, software artifacts, tutorials, online courses, workshops, public talks, or open repositories. 
This practice ensures that the knowledge generated during project execution is preserved and disseminated beyond the immediate project team. Systematic documentation also prevents knowledge loss as senior members graduate and facilitates the onboarding of new participants, contributing to the long-term sustainability of the league, as discussed later in the paper.

\section{Representative Projects}\label{sec:projects}

The organizational framework presented in the previous section has been refined through the development of projects addressing different educational, scientific, and extension objectives. Although these initiatives differ in scope and technical focus, they all follow the same organizational principles, combining student protagonism, project-based learning, and the integration of teaching, research, and extension.

This section presents representative projects developed within LIA to illustrate the proposed framework in practice. Rather than providing an exhaustive description of all initiatives conducted by the league, the selected projects demonstrate how the organizational model supports different types of activities, including competitions, scientific dissemination, open educational resources, study groups, and community engagement.

\subsection{Competition Teams}


Competition-oriented projects constitute one of the most effective mechanisms for promoting project-based learning in Artificial Intelligence. Besides requiring students to apply theoretical concepts to solve complex problems, competitions encourage teamwork, software engineering practices, autonomous learning, and decision-making under realistic constraints. Within the proposed organizational framework, competition teams naturally integrate teaching, research, and extension, while providing students with opportunities to develop both technical and transversal competencies.

One of the first and most successful competition teams created within LIA was the LI(A)RA project, dedicated to the development of intelligent agents for the Multi-Agent Programming Contest (MAPC)\footnote{\url{https://multiagentcontest.org/}}. The project also served as one of the first validations of the organizational framework adopted by the league, demonstrating how competition-based initiatives could be systematically integrated into undergraduate education. Table~\ref{tab:liara} summarizes the main characteristics of the project, which are discussed in the following paragraphs.

The project originated from the undergraduate course ``Specialized Topics I'' (STI), offered within the Computer Engineering curriculum. Rather than separating classroom activities from extracurricular projects, the competition team was created as a natural extension of the course. The theoretical concepts presented during the lectures provided students with the necessary background in multi-agent systems, while project activities enabled them to apply this knowledge to a realistic and open-ended engineering problem.

During the initial phase of the project, students were organized into study groups responsible for investigating previous editions of the competition, including publications describing the winning solutions~\cite{ahlbrecht2022multi,ahlbrecht2021multi}. The groups analyzed existing strategies, software architectures, infrastructure requirements, and engineering practices adopted by previous participants. This collaborative investigation established a common technical foundation for the team while promoting critical reading, technical discussions, and knowledge sharing among students.

The knowledge acquired during the study phase was transformed into extension products. Students documented the concepts, methodologies, and implementation techniques in a booklet intended to disseminate the acquired knowledge beyond the project itself. Based on this material, the team developed a self-instructional online course composed of recorded lectures, implementation examples, practical exercises, quizzes, and assessment activities. The course continues to be offered by UFSC and, as of 2025, has completed four editions, extending the educational impact of the original competition project to a much broader audience.

After the preparation phase, the team implemented a complete solution for the competition and participated in the MAPC, achieving fourth place in the final competition. Beyond the competitive result, participation enabled students to exchange experiences with teams from other institutions, compare alternative engineering solutions, and evaluate their own approaches in an international environment. The developed solution and its design decisions were documented in the form of a book chapter~\cite{custodio2022li}, further contributing to the dissemination and reproducibility of the project.

Beyond its technical outcomes, the LI(A)RA project demonstrated the educational potential of the proposed organizational framework. Students assumed responsibility for planning activities, organizing study groups, coordinating development tasks, and making technical decisions, while faculty members acted as mentors throughout the process. The project also validated several organizational practices that were subsequently adopted by other initiatives within LIA, including the systematic use of study groups, the transformation of project outcomes into educational resources, and the integration of teaching, research, and extension within a single student-centered initiative. The study group structure originally established in LI(A)RA has since been replicated in new initiatives focused on agent technologies and has also served as the foundation for the GAIA project described later in this paper, demonstrating the framework's adaptability across different application domains.

\begin{table}[t]
\centering
\renewcommand{\arraystretch}{1.2}
\begin{tabular}{p{2.7cm}|p{12.3cm}}
\hline
\textbf{Aspect} & \textbf{Description} \\
\hline \hline
Primary Goal &
Develop intelligent agents for the Multi-Agent Programming Contest (MAPC) while providing students with a project-based learning experience in multi-agent systems. \\
\hline
Teaching &
Integrated with the undergraduate course ``Specialized Topics I'' (STI), where theoretical concepts were complemented by study groups and collaborative learning activities. \\
\hline
Research &
Students investigated previous MAPC editions, analyzed state-of-the-art strategies and software architectures, and documented the developed solution in a book chapter. \\
\hline
Extension &
The acquired knowledge was transformed into a booklet and a self-instructional online course, disseminating the project outcomes to students and the broader community. \\
\hline
Student Development &
Students organized study groups, coordinated technical activities, made engineering decisions, and developed leadership, teamwork, software engineering, and project management skills. \\
\hline
Main Outcomes &
Fourth place in the MAPC final competition, publication of a book chapter, creation of educational materials, and establishment of organizational practices later adopted by other LIA projects. \\
\hline \hline
\end{tabular}
\caption{Summary of the LI(A)RA project according to the proposed organizational framework}
\label{tab:liara}
\end{table}

\subsection{Open Talks about Artificial Intelligence} 


Another representative initiative developed within LIA is the ``Open Talks about Artificial Intelligence'', an annual cycle of lectures dedicated to the dissemination, discussion, and critical reflection on topics related to Artificial Intelligence. Unlike competition-oriented projects, whose primary objective is the development of technical solutions, the Open Talks were conceived as a permanent extension initiative aimed at promoting dialogue between the university and the broader community. Table~\ref{tab:opentalks} provides an overview of the project, while the following paragraphs describe its main activities and outcomes in greater detail.

The talks are open to undergraduate and graduate students, researchers, industry professionals, educators, and AI enthusiasts, fostering an inclusive and multidisciplinary environment for knowledge exchange. Speakers include faculty members, invited researchers, professionals from industry, and students presenting the results of their own projects. All talks are publicly available through LIA's YouTube channel\footnote{\url{https://www.youtube.com/watch?v=qXvQLzxgOf4&list=PL-ojBVkz9Su9NIn_XMls4vxIWe7Ma9PHn}}, enabling the generated content to remain accessible beyond the event itself.

From the teaching perspective, the Open Talks are integrated with undergraduate and graduate courses related to Artificial Intelligence. Several lectures are aligned with topics covered in the curriculum and are incorporated into classroom activities as complementary learning resources. Students are encouraged to discuss the presented topics, relate them to theoretical concepts introduced during lectures, and reflect on current research challenges and industrial applications.

The initiative also promotes the integration of research activities. Undergraduate and graduate students are encouraged to present the outcomes of scientific initiation projects, course assignments, master's dissertations, and ongoing research developed within LIA and collaborating laboratories. These presentations provide students with valuable experience in scientific communication while encouraging interaction between research groups and the broader academic community.

From the extension perspective, the Open Talks constitute a continuous outreach activity through which contemporary AI knowledge is shared with society. By bringing together participants from academia, industry, and the general public, the initiative promotes discussions on both technical and societal aspects of Artificial Intelligence, including machine learning, computer vision, multi-agent systems, AI ethics, healthcare applications, and emerging industrial technologies. This interaction contributes to democratizing access to AI knowledge while strengthening the relationship between the university and external communities.

Beyond the technical content of the lectures, the project also plays an important organizational role within LIA. Students are responsible for identifying relevant themes, inviting speakers, organizing the event schedule, managing registrations, promoting the talks through social media, moderating discussions, and documenting each edition. These responsibilities provide opportunities for students to develop leadership, communication, teamwork, and project management skills that complement their technical education.

The Open Talks is a good example of how a relatively simple initiative can integrate teaching, research, and extension. Teaching is strengthened through the incorporation of lectures into formal courses, research is disseminated through student and faculty presentations, and extension is achieved by providing free and permanent public access to high-quality educational content. At the same time, the organizational responsibilities assumed by students reinforce the student-centered philosophy that characterizes the organizational framework.

\begin{table}[ht]
\centering
\renewcommand{\arraystretch}{1.2}
\begin{tabular}{p{2.7cm}|p{12.3cm}}
\hline
\textbf{Aspect} & \textbf{Description} \\
\hline \hline
Primary Goal &
Disseminate Artificial Intelligence knowledge through open lectures while promoting interaction between academia, industry, and society. \\
\hline
Teaching &
Lectures are integrated into undergraduate and graduate courses as complementary learning resources, connecting theoretical concepts with contemporary AI applications and research. \\
\hline
Research &
Students, researchers, and invited speakers present scientific results and ongoing research, fostering scientific communication and interdisciplinary collaboration. \\
\hline
Extension &
The talks are open to the external community and permanently available through LIA's YouTube channel, providing free and continuous access to AI educational content. \\
\hline
Student Development &
Students organize the entire event, including speaker invitations, publicity, scheduling, moderation, recording, and dissemination, developing leadership, communication, teamwork, and organizational skills. \\
\hline
Main Outcomes &
Annual lecture series, publicly available recorded talks, strengthened interaction between academia and society, and continuous dissemination of AI knowledge. \\
\hline \hline
\end{tabular}
\caption{Main aspects of the Open Talks about Artificial Intelligence in the context of the proposed organizational framework.}
\label{tab:opentalks}
\end{table}

\subsection{Repository of Knowledge in Artificial Intelligence}


The Repository of Knowledge in Artificial Intelligence (RepoAI) is a permanent extension initiative developed within LIA to consolidate, organize, and disseminate educational and technical resources related to Artificial Intelligence. Unlike the previous projects, whose primary outcomes are competitions or public events, RepoAI functions as a continuously evolving knowledge repository, ensuring that the technical knowledge generated within the league is documented and made publicly available. We summarize the key aspects of this project in Table~\ref{tab:repoai} and detail them as follows.

RepoAI consists of an open repository\footnote{\url{https://github.com/Liga-IA/RepoAI}} containing tutorials, implementation examples, educational materials, and software artifacts covering diverse topics in Artificial Intelligence. The repository is collaboratively maintained by students and faculty members, allowing new materials to be incorporated as projects evolve and new technologies emerge.

From the teaching perspective, RepoAI serves as a complementary educational resource for undergraduate and graduate courses. Students are encouraged to transform course assignments, study group reports, and project deliverables into structured tutorials, requiring them to reorganize acquired knowledge into clear and reusable educational materials. This process reinforces students' understanding of the studied concepts while producing resources that support future learning activities.

RepoAI also establishes a direct connection with research activities. Tutorials frequently originate from scientific initiation projects, graduate research, competition teams, and study groups, where students investigate specific Artificial Intelligence techniques in greater depth. By documenting methodologies, implementation details, and experimental procedures, research outcomes become reusable educational artifacts that facilitate knowledge transfer both within and beyond the university.

From the extension perspective, RepoAI provides unrestricted public access to educational resources, reducing barriers for students, educators, professionals, and AI enthusiasts interested in learning Artificial Intelligence. Unlike traditional scientific publications, the repository emphasizes practical implementation, reproducibility, and progressive learning, combining conceptual explanations with executable code examples and implementation guidelines.

The repository also plays a strategic organizational role within LIA. Students participating in different projects are encouraged to document their work as a final project deliverable, ensuring that the knowledge produced is preserved after the project's conclusion. Consequently, RepoAI has become the primary institutional repository for educational materials generated by the league, supporting the onboarding of new members while avoiding the loss of accumulated knowledge between successive student generations.

RepoAI illustrates how digital knowledge repositories can support teaching, research, and extension. Educational materials produced during project execution become reusable learning resources, research outcomes gain broader visibility through practical tutorials, and extension is realized through open and permanent public access to AI knowledge. Furthermore, the repository establishes a mechanism for preserving the institutional memory of the league, allowing knowledge generated by one generation of students to be reused and expanded by the next.

\begin{table}[ht]
\centering
\renewcommand{\arraystretch}{1.2}
\begin{tabular}{p{2.7cm}|p{12.3cm}}
\textbf{Aspect} & \textbf{Description} \\
\hline \hline
Primary Goal &
Preserve, organize, and disseminate Artificial Intelligence knowledge through an open repository of educational materials and implementation examples. \\
\hline
Teaching &
Students transform course assignments, study group reports, and project outcomes into tutorials and educational resources that support undergraduate and graduate courses. \\
\hline
Research &
Research outcomes, scientific initiation projects, competition teams, and study groups are documented as reproducible tutorials and software artifacts, facilitating knowledge transfer and reuse. \\
\hline
Extension &
The repository is openly available to students, educators, professionals, and the general public, providing unrestricted access to educational materials and AI implementation examples. \\
\hline
Student Development &
Students develop technical writing, scientific communication, documentation, software engineering, and collaborative development skills while producing reusable educational resources. \\
\hline
Main Outcomes &
Open repository of tutorials and code examples, permanent educational resources, preservation of institutional knowledge, and continuous support for the onboarding and training of new LIA members. \\
\hline \hline
\end{tabular}
\caption{Summary of the RepoAI project according to the proposed organizational framework}
\label{tab:repoai}
\end{table}

\subsection{Study Groups}

Study groups constitute one of the fundamental educational mechanisms of the proposed organizational framework. Unlike the other initiatives described in this section, study groups are not independent projects but rather a collaborative learning strategy that supports every activity developed within LIA. They provide the environment in which students acquire the theoretical and technical background required before engaging in research, software development, competitions, or extension activities. Table~\ref{tab:studygroups} summarizes the key aspects of this project.

Study groups are created during the initial phase of a project or whenever members identify the need to investigate a new topic in greater depth. Each group is organized around a specific theme, such as machine learning, computer vision, multi-agent systems, large language models, or software engineering practices. Students collaboratively select reference materials, scientific papers, textbooks, technical documentation, and software libraries, which are then discussed through regular meetings and seminars.

\begin{table}[ht]
\centering
\renewcommand{\arraystretch}{1.2}
\begin{tabular}{p{2.7cm}|p{12.3cm}}\hline
\textbf{Aspect} & \textbf{Description} \\
\hline \hline
Primary Goal &
Provide a collaborative learning environment that supports the development of all projects within LIA through the study of emerging topics and state-of-the-art technologies. \\
\hline
Teaching &
Complement undergraduate courses through collaborative learning, seminars, practical demonstrations, and the exploration of topics beyond the formal curriculum. \\
\hline
Research &
Support literature reviews, identification of research questions, prototype development, and the initiation of scientific projects, competition teams, and graduate research. \\
\hline
Extension &
Transform study materials into tutorials, presentations, online courses, recorded lectures, and educational resources incorporated into other LIA extension initiatives. \\
\hline
Student Development &
Develop scientific reading, critical thinking, technical communication, collaborative learning, leadership, and presentation skills through peer-led activities. \\
\hline
Main Outcomes &
Technical reports, tutorials, educational materials, research ideas, software prototypes, and knowledge that supports competitions, research projects, and extension initiatives. \\
\hline \hline
\end{tabular}
\caption{Summary of the Study Groups projects according to the proposed organizational framework}
\label{tab:studygroups}
\end{table}

Rather than relying exclusively on faculty-led instruction, study groups are coordinated by students, usually those with previous experience in the subject. Members are encouraged to alternate responsibilities for selecting reading materials, preparing presentations, leading technical discussions, and conducting practical demonstrations. This collaborative organization promotes active learning while strengthening communication, critical thinking, and scientific argumentation skills.

From the teaching perspective, study groups complement the formal curriculum by allowing students to explore topics that are not fully covered in undergraduate courses or to deepen their understanding of concepts introduced in the classroom. Since participation is voluntary and driven by students' interests, the groups also provide flexibility to address emerging technologies and research trends.

Study groups also establish the foundation for research activities. The collaborative investigation of the scientific literature frequently results in research questions, software prototypes, and experimental evaluations that subsequently evolve into scientific initiation projects, undergraduate theses, master's dissertations, competition teams, and research collaborations. In this sense, study groups often represent the first stage of the research process within LIA.

The outcomes of study groups are transformed into extension products whenever possible. Technical reports, tutorials, presentations, recorded lectures, software examples, and educational materials produced during the meetings are incorporated into initiatives such as RepoAI, the Open Talks about Artificial Intelligence, or online courses. Thus, the knowledge acquired by one group becomes available to future students and to the broader community, reinforcing the cumulative nature of the league's educational activities.

More important, study groups represent the primary mechanism through which the proposed organizational framework integrates teaching, research, and extension. By transforming collaborative learning into research outcomes and into publicly available educational resources, study groups establish a continuous process of knowledge generation, dissemination, and reuse that supports every initiative developed within LIA.

\subsection{Accessible Educational Technologies}

Beyond initiatives focused on competitions, knowledge dissemination, and collaborative learning, LIA also develops extension projects aimed at producing social relevant technologies. These projects are typically carried out in collaboration with external members of the community, allowing students to apply Artificial Intelligence techniques to address real problems, including educational and accessibility challenges, while strengthening the connection between the university and society.

One representative example is the GAIA (Games and Artificial Intelligence Applications) project, an extension initiative dedicated to the development of interactive applications based on Artificial Intelligence and Computer Vision. The project explores vision-based interaction to develop AI-powered interactive applications with social impact, combining human-computer interaction with solutions for education, accessibility, rehabilitation, and health. In addition to serving as a platform for research in computer vision and intelligent interfaces, GAIA provides students with opportunities to investigate accessibility, educational technologies, human-computer interaction, and serious games within a multidisciplinary environment. Table~\ref{tab:gaia} summarizes this project, which is detailed as follow. 

\begin{table}[t]
\centering
\renewcommand{\arraystretch}{1.2}
\begin{tabular}{p{2.7cm}|p{12.3cm}}\hline
\textbf{Aspect} & \textbf{Description} \\
\hline \hline
Primary Goal &
Develop AI-powered interactive applications with social impact by combining Computer Vision, natural human-computer interaction, and intelligent systems to address challenges in education, accessibility, rehabilitation, and healthcare. \\
\hline
Teaching &
Students apply concepts from Artificial Intelligence, Computer Vision, software engineering, human-computer interaction, and accessibility through the development of multidisciplinary applications. \\
\hline
Research &
Investigation of gesture recognition, computer vision, interactive systems, serious games, accessibility, rehabilitation technologies, and AI-assisted educational applications, resulting in scientific publications and software artifacts. \\
\hline
Extension &
Development and public demonstration of interactive applications for education, accessibility, rehabilitation, and healthcare during outreach activities, school visits, technology exhibitions, and community events. \\
\hline
Student Development &
Software engineering, computer vision, AI application development, interdisciplinary collaboration, accessibility-aware design, user-centered development, and science communication. \\
\hline
Main Outcomes &
Interactive applications with social impact, including \textit{Forca Libras}, AI-based rehabilitation prototypes, scientific publications, software artifacts, and outreach demonstrations promoting accessibility and digital inclusion. \\
\hline \hline
\end{tabular}
\caption{Summary of the GAIA project according to the proposed organizational framework}
\label{tab:gaia}
\end{table}

Among the educational applications developed within the project, \textit{Forca Libras}\footnote{\url{https://forcalibras.ufsc.br/}} (The hangman in LIBRAS) represents one of its most significant outcomes. The application employs computer vision techniques to recognize hand gestures corresponding to the Brazilian Sign Language (LIBRAS), enabling users to interact with the game while practicing the manual alphabet. Unlike traditional educational games, \textit{Forca Libras} adopts a dual pedagogical approach. For hearing students, it functions as a tool for learning and practicing LIBRAS through gesture recognition. Further, for deaf students, the game serves as a pedagogical resource to support Portuguese literacy by combining visual representations, contextual images, textual hints, and Portuguese words. This association between signs, images, and written language creates an inclusive learning environment that supports second-language acquisition and accessible education.

The development of \textit{Forca Libras} required students to integrate concepts from Artificial Intelligence, Computer Vision, software engineering, educational game design, and accessibility. Beyond the technical implementation, students were involved in literature review, interface design, experimental evaluation, and the preparation of scientific publications, reinforcing the project's strong connection with undergraduate education and research. The resulting system was published at the Brazilian Symposium on Collaborative Systems (SBSC) 2026, demonstrating the potential of AI-based interactive educational technologies to support inclusive learning environments~\cite{forcalibras-2026}.

From the extension perspective, GAIA has also become an important instrument for science outreach. The developed applications are frequently demonstrated during university extension events, including visits from elementary and high school students, technology exhibitions, and activities that introduce the community to Artificial Intelligence. During these events, LIA members present the interactive games while discussing the role of AI in education, accessibility, and digital inclusion, helping to demystify Artificial Intelligence and encouraging young students to pursue careers in computing and engineering.

Overall, the GAIA project illustrates how the proposed organizational framework can generate extension initiatives with direct societal impact. By integrating Artificial Intelligence, accessibility, and inclusive education within a single project, GAIA demonstrates that student-led initiatives can produce scientific knowledge, innovative educational technologies, and meaningful contributions to the community, reinforcing the continuous integration of teaching, research, and extension that characterizes the LIA framework.

\subsection{Other Extension Initiatives}

In addition to the representative projects discussed in the previous sections, LIA develops other extension initiatives that emerge from different contexts, including new research opportunities, institutional collaborations, and community demands. The dynamic nature of the proposed organizational framework allows new projects to be incorporated into the league while preserving the same student-centered philosophy and the integration of teaching, research, and extension.

One recurring activity consists of supporting university outreach events, in particular the project ``\textit{Visitas Guiadas}'' (Guided visits) that receive students from elementary and high schools, introducing them to the university environment and to current research in Artificial Intelligence. During these visits, LIA members organize laboratory demonstrations, present ongoing research projects, and promote interactive activities designed to make AI concepts accessible to younger audiences. These initiatives not only strengthen the relationship between the university and society but also encourage prospective students to pursue careers in computing and engineering.


It is important to emphasize that the league also develops projects beyond the field of Artificial Intelligence. One such initiative is Offensive Security in Practice (OffSec), which aims to teach computer security from the attacker's perspective. To achieve this objective, the project combines hands-on challenges with student-maintained knowledge repositories. The project is open to both university students and external participants, who develop offensive security skills by solving challenges in simulated Capture the Flag (CTF) competitions, and documenting the acquired knowledge through tutorials maintained in the project's repository. More experienced students curate both the educational material and the CTF challenges, ensuring that participants progressively acquire the knowledge required to solve increasingly challenging CTF problems. In addition to these internal activities, participants prepare and deliver outreach lectures, offered both online and at local schools, thereby extending the project's educational impact to the broader community.

These examples illustrate an important characteristic of the proposed framework: projects are not static but continuously evolve according to new opportunities, technological advances, and societal demands. While individual initiatives may differ in their objectives and target audiences, they all follow the same organizational principles, reinforcing the flexibility and sustainability of the proposed model.

\section{Lessons Learned}\label{sec:lessons}

The representative projects presented in the previous section demonstrate the flexibility of the proposed organizational framework. Although each initiative was created to address distinct educational, scientific, or outreach objectives, they all share a common organizational philosophy centered on student protagonism, project-based learning, and the continuous integration of teaching, research, and extension.
Rather than operating as isolated initiatives, the projects collectively contribute to complementary organizational dimensions, including collaborative learning, project-based development, knowledge management, scientific dissemination, leadership development, and the integration of the three university pillars, as summarized in Table~\ref{tab:framework_dimensions}. 
This perspective highlights that the educational value of the framework emerges not only from the individual projects themselves but also from the interactions among them, which together form a coherent and sustainable educational ecosystem. 

The following subsections discuss the lessons learned throughout the implementation of the proposed framework and its instantiation towards the Academic League of Artificial Intelligence (LIA).

\begin{table}[t]
\centering
\renewcommand{\arraystretch}{1.15}
\small
\begin{tabular}{p{3.4cm}|p{2.8cm}|p{8.5cm}}
\hline
\textbf{Framework Dimension} & \textbf{Representative Projects} & \textbf{Contribution} \\
\hline \hline
Knowledge Acquisition & Study Groups, LI(A)RA, GAIA & Develop technical and scientific competencies through collaborative learning, literature reviews, experimentation, and project-based activities. \\
\hline
Project Development & LI(A)RA, GAIA & Transform acquired knowledge into software artifacts, research prototypes, educational technologies, and scientific outcomes through student-led projects. \\
\hline
Knowledge Preservation & RepoAI, Study Groups, LI(A)RA & Document tutorials, software artifacts, reports, and educational materials to preserve institutional knowledge and support the onboarding of new members. \\
\hline
Knowledge Dissemination & Open Talks, RepoAI, LI(A)RA, GAIA & Disseminate knowledge through public lectures, online repositories, educational resources, publications, outreach activities, and demonstrations. \\
\hline
Integration of Teaching, Research, and Extension & All projects & Each initiative contributes to one or more university pillars, while the portfolio as a whole ensures their continuous integration within a student-centered educational ecosystem. \\
\hline
\end{tabular}
\caption{Organizational dimensions of the proposed framework and representative projects contributing to each dimension.}
\label{tab:framework_dimensions}
\end{table}

\subsection{Student Protagonism as the Driving Force}

One of the defining characteristics of the proposed framework is its student-centered organization. Unlike traditional extracurricular activities, where students often participate as collaborators in faculty-led projects, the initiatives developed within LIA are primarily conceived, organized, and executed by students. Faculty advisors provide academic guidance, institutional support, and technical mentoring, while operational and strategic decisions are delegated to the administrative board and project leaders.

This organizational model encourages students to progressively assume greater responsibilities throughout their participation in the league. New members initially contribute to ongoing activities, subsequently coordinate individual tasks, and eventually assume leadership positions within projects or the administrative board. Since both administrative positions and project leadership are democratically elected by members, leadership naturally emerges from technical competence, commitment, and previous experience rather than being assigned by faculty members.

Beyond technical education, this structure contributes to the development of transversal competencies that are valued in engineering and computing education. Throughout the reported projects, students were responsible for organizing teams, managing schedules, communicating with external partners, preparing educational resources, coordinating extension activities, and making technical decisions. Consequently, the league functions not only as a technical learning environment but also as a space for leadership development, collaborative decision-making, and professional growth.

\subsection{Integrating Teaching, Research, and Extension Through Projects}

The proposed framework adopts projects as the primary mechanism for integrating the three constitutional pillars of higher education. Rather than treating teaching, research, and extension as independent institutional activities, projects naturally evolve across these dimensions according to their objectives and maturity.


The representative projects presented in Section~\ref{sec:projects} emphasize different aspects of the three university pillars while collectively providing a balanced educational ecosystem. LI(A)RA integrated teaching and research while generating extension outcomes such as online courses and educational materials. Open Talks strengthen extension by disseminating scientific knowledge to the broader community while supporting undergraduate education. RepoAI contributes equally to teaching, research, and extension by preserving and disseminating the knowledge generated across multiple initiatives. Similarly, Study Groups establish the educational foundation from which many research and extension activities emerge. Finally, GAIA demonstrates how multidisciplinary AI projects can integrate teaching, research, and extension through the development of interactive applications with social impact in domains such as education, accessibility, and healthcare.

These observations indicate that the integration of teaching, research, and extension does not require every project to contribute equally to each dimension. Instead, the proposed framework relies on a complementary portfolio of initiatives whose combined contributions operationalize the inseparability of the three university pillars. This flexibility enables the league to adapt its portfolio according to emerging opportunities while maintaining its educational objectives.

\subsection{Knowledge Management and Organizational Sustainability}

Student organizations naturally face the challenge of maintaining continuity despite the periodic graduation of their members. Without appropriate knowledge management practices, technical expertise and organizational experience may be lost, requiring successive generations to recreate previous work.

The proposed framework addresses this challenge by transforming project outcomes into persistent educational artifacts. Tutorials, software repositories, technical reports, online courses, recorded lectures, scientific publications, and other educational resources preserve the knowledge generated throughout project execution while facilitating the onboarding of new members. In particular, RepoAI has become the central repository for documenting and organizing educational resources produced by the league, while initiatives such as Open Talks and competition projects continuously contribute new content.

This strategy establishes a cumulative learning process in which projects no longer represent isolated experiences but instead become part of the institutional memory of the league. As new students join LIA, they build upon the knowledge generated by previous cohorts, allowing projects to progressively increase in complexity and maturity rather than restarting from the beginning.

\subsection{Replicability and Limitations}

Although the proposed framework was developed within an academic league dedicated to Artificial Intelligence, its organizational principles are largely independent of the technical domain. The separation between administrative governance and project execution, the democratic election of leadership positions, the student-centered management philosophy, and the systematic transformation of project outcomes into educational and extension products can be adapted to academic leagues in other areas of computing and engineering, including robotics, software engineering, cybersecurity, embedded systems, and data science.

Another important characteristic of the framework is its accessibility. None of the representative projects required large financial investments or specialized infrastructure to be implemented. Instead, their success primarily depended on student engagement, faculty mentorship, collaborative organization, and the effective integration of teaching, research, and extension. Consequently, the framework may be adopted by institutions with different sizes, resources, and educational contexts.

Nevertheless, this work also presents some limitations. The reported experiences originate from a single academic league at a single institution, and the evaluation is primarily qualitative, focusing on the organizational aspects of the proposed framework. Although the representative projects demonstrate the feasibility of the model, broader empirical studies involving multiple institutions and quantitative assessments of student learning outcomes would provide stronger evidence regarding its educational impact and generalizability.

\section{Conclusion}\label{sec:conclusion}

This paper presented a student-centered organizational framework for academic leagues in Computing and Engineering, using the Academic League of Artificial Intelligence (LIA) at the Federal University of Santa Catarina as a representative case study. Motivated by the lack of documented organizational models for academic leagues in these fields, the proposed framework demonstrates how teaching, research, and university extension can be systematically integrated through complementary, project-based initiatives under a stable governance structure.

The reported experience shows that the educational value of the framework emerges not only from individual projects but also from their interaction as part of a coherent educational ecosystem. The representative initiatives discussed in this paper illustrate how competition teams, study groups, knowledge repositories, scientific dissemination activities, and socially oriented AI applications collectively promote technical education, scientific production, leadership development, knowledge preservation, and community engagement. In particular, the combination of democratic governance, student protagonism, systematic knowledge management, and project-based learning has contributed to the long-term sustainability of the league while enabling successive generations of students to build upon previous experiences.

Although the framework was instantiated within an academic league dedicated to Artificial Intelligence, its organizational principles are largely independent of the technical domain and may be adapted to other areas of Computing and Engineering. By documenting both the organizational structure and the lessons learned throughout its implementation, this work provides a practical reference for institutions interested in establishing or strengthening student-centered academic leagues and similar educational organizations.

Future work includes evaluating the framework across multiple institutions and academic leagues, as well as conducting quantitative studies to assess its impact on student learning, leadership development, scientific production, retention, and engagement in university extension activities. We also expect to investigate how the proposed framework can be adapted to emerging educational contexts, interdisciplinary initiatives, and international collaborations in Computing and Engineering education.

\section*{Acknowledgments}

The authors acknowledge the use of an AI assistant to review the manuscript for grammar and formal writing style. The authors are responsible for the content, interpretations, and conclusions presented in this paper.

\bibliographystyle{splncs04}
\bibliography{references}

@article{goergen2023exploratory,
  title={An exploratory study of the academic leagues in southern Brazil: doing multiple activities},
  author={Goergen, Diego In{\'a}cio and Antonello, Ivan Carlos Ferreira and Costa, Bartira Erc{\'\i}lia Pinheiro da},
  journal={Revista Brasileira de Educa{\c{c}}{\~a}o M{\'e}dica},
  volume={47},
  number={01},
  pages={e12},
  year={2023},
  publisher={SciELO Brasil}
}

@ARTICLE{Fernandes2025,
	author = {Fernandes, Isabelle Marques and Casari, Julia Ravazzi and Gonsalves, Daniel Gregório and Neto, Wilson Falco and Rissi, Renato and Pedroza Cavalcante, Ana Suelen},
	title = {After all, what is an Academic Health League? Analysis considering a Systematic Review; [Afinal, o que é uma Liga Acadêmica em Saúde? Análise à luz de uma Revisão Sistemática]},
	year = {2025},
	journal = {Medicina (Brazil)},
	volume = {58},
	number = {4},
	doi = {10.11606/issn.2176-7262.rmrp.2025.228101}
}

@ARTICLE{Camilo2020,
	author = {Camilo, Gustavo Bittencourt and Bastos, Marcus Gomes and Toledo, Gabriela Cumani and Ferreira, Ana Paula and Brandão, Tainá Gomes and Miranda Reis, Ana Flávia and de Almeida Paes Barretto Coutinho, Isabella and Aranha, Gabriel Lunardi and deSouza, Bárbara Isadora Amâncio},
	title = {Analysis of academic medical leagues from the students’ perspective; [Análise das ligas acadêmicas de medicina sob a perspectiva dos alunos]},
	year = {2020},
	journal = {Scientia Medica},
	volume = {30},
	number = {1},
	doi = {10.15448/1980-6108.2020.1.36190},
	type = {Article},
	publication_stage = {Final},
	source = {Scopus},
    }

@ARTICLE{DosSantosSilva2021,
	author = {Dos Santos Silva, John Victor and Dos Santos Júnior, Claudio José and Dos Santos, Larissa Dandara Lima and Da Silva Barbosa, Vívian Mayara and Brandão, Thyara Maia and Ribeiro, Mara Cristina},
	title = {Interdisciplinary academic league of mental health: Expanding the formation and the practice in the field of psychosocial care},
	year = {2021},
	journal = {Medicina (Brazil)},
	volume = {54},
	number = {2},
	doi = {10.11606/issn.2176-7262.rmrp.2021.174130},
	type = {Article},
	publication_stage = {Final},
	source = {Scopus},
}

@misc{matheus2019important,
  title={The important role of academic leagues (extensions) in Brazilian medical education},
  author={Matheus, Belloni Torsani},
  journal={Revista da Associa{\c{c}}{\~a}o M{\'e}dica Brasileira},
  volume={65},
  pages={98--99},
  year={2019},
  publisher={SciELO Brasil}
}

@article{santos2020strengthening,
  title={Strengthening teaching, research and university extension through Academic Leagues},
  author={Santos, Fernanda Batista Oliveira and dos Santos Carregal, Fernanda Alves and Schreck, Rafaela Siqueira Costa and Diniz, Thiago Frederico and Siman, Andr{\'e}ia Guerra and Braga, Luciene Muniz and Matozinhos, Fernanda Penido and Barbosa, Jaqueline Almeida Guimar{\~a}es},
  journal={Brazilian Journal of Health Review},
  volume={3},
  number={2},
  pages={3439--3447},
  year={2020}
}

@article{cavalcante2021search,
  title={In search of the contemporary definition of “academic leagues” based on the health sciences’ experience},
  author={Cavalcante, Ana Suelen Pedroza and Vasconcelos, Maristela In{\^e}s Osawa and Ceccim, Ricardo Burg and Maciel, Gabriel Pereira and Ribeiro, Marcos Aguiar and Henriques, Regina Lucia Monteiro and Albuquerque, Izabelle Napole{\~a}o Mont’Alverne and Silva, Maria Rocineide Ferreira da},
  journal={Interface-Comunica{\c{c}}{\~a}o, Sa{\'u}de, Educa{\c{c}}{\~a}o},
  volume={25},
  pages={e190857},
  year={2021},
  publisher={SciELO Public Health}
}

@article{oliveira2024transformative,
  title={The transformative integration of university extension and education in communities},
  author={Oliveira, DA dos S and Bernet, RR and Hoyos, DC de M},
  journal={Seven Editora},
  volume={576},
  pages={82},
  year={2024}
}

@article{gimenez2024three,
  title={The Three Fundamental University Missions: Brazilian Challenges},
  author={Gimenez, Ana Maria Nunes and de Oliveira Gavira, Muriel and Bonacelli, Maria Beatriz M},
  journal={International Higher Education},
  number={120},
  year={2024}
}

@article{Guimaraes2024, 
title={The Higher Education Triad: Research, Extension, and Teaching as Pillars of Academic Formation }, volume={21}, 
url={https://periodicos.unis.edu.br/mythos/article/view/912}, 
DOI={10.36674/mythos.v21i2.912}, 
number={2}, 
journal={Revista Mythos}, 
author={Guimarães Júnior, Ernani de Souza},
year={2024}, 
month={Sep.}, 
pages={100–104} 
}

@article{de2024context,
  title={CONTEXT BETWEEN UNIVERSITY EXTENSION AND CONTINUING TEACHER TRAINING: CHALLENGES AND POSSIBILITIES},
  author={de Oliveira Brito, Renato and Chesini, Cl{\'a}udia and de Lucena, Jos{\'e} Ivaldo Ara{\'u}jo and others},
  journal={Journal of Media Critiques},
  volume={10},
  number={26},
  pages={e56--e56},
  year={2024}
}

@inproceedings{souza2022third,
  title={THE THIRD MISSION IN BRAZIL AND THE LINK BETWEEN TEACHING AND RESEARCH},
  author={Souza, D and Mendon{\c{c}}a, J},
  booktitle={INTED2022 Proceedings},
  pages={9789--9793},
  year={2022},
  organization={IATED}
}

@incollection{custodio2022li,
  title={LI(A)RA Team-A Declarative and Distributed Implementation for the MAPC 2022},
  author={Cust{\'o}dio, Marcelo and Rocha, Michele and Battaglin, Ricardo and Farias, Giovani P and Panisson, Alison R},
  booktitle={Multi-Agent Progamming Contest},
  pages={165--194},
  year={2022},
  publisher={Springer}
}

@incollection{ahlbrecht2022multi,
  title={The multi-agent programming contest 2022},
  author={Ahlbrecht, Tobias and Dix, J{\"u}rgen and Fiekas, Niklas and Krausburg, Tabajara},
  booktitle={Multi-Agent Progamming Contest},
  pages={1--18},
  year={2022},
  publisher={Springer}
}

@book{ahlbrecht2021multi,
  title={Multi-Agent Programming Contest 2021},
  author={Ahlbrecht, Tobias and Dix, J{\"u}rgen and Fiekas, Niklas and Krausburg, Tabajara},
  year={2021},
  publisher={Springer}
}

@inproceedings{forcalibras-2026,
 author = {Italo da Silva and João José Sebastião and Ruan Pablo Soares and Beatriz Tartare and Gabriel Santos and Maria Eduarda Vianna and Sheila Oliveira and Talita Caneda and Alison Panisson},
 title = { An Inclusive AI-Based Game for LIBRAS Learning Developed Through University Extension},
 booktitle = {Anais do XXI Simpósio Brasileiro de Sistemas Colaborativos},
 location = {Porto Alegre/RS},
 year = {2026},
 issn = {2326-2842},
 pages = {327--340},
 publisher = {SBC},
 address = {Porto Alegre, RS, Brasil},
 doi = {10.5753/sbsc.2026.20041},
 url = {https://sol.sbc.org.br/index.php/sbsc/article/view/42952}
}

@misc{brasil1988constituicao,
  title = {Constituição da República Federativa do Brasil},
  author = {{Brasil}},
  year = {1988},
  howpublished = {Brasília, DF: Presidência da República},
  url = {https://www.planalto.gov.br/ccivil_03/constituicao/constituicaocompilado.htm},
  note = {Artigo 207}
  }

@misc{forproex2012,
  author       = {{Fórum de Pró-Reitores de Extensão das Instituições Públicas de Educação Superior Brasileiras (FORPROEX)}},
  title        = {Política Nacional de Extensão Universitária},
  year         = {2012},
  address      = {Manaus, Brazil},
  publisher    = {FORPROEX},
  url          = {https://www.ufmg.br/proex/renex/images/documentos/2012-07-13-Politica-Nacional-de-Extensao.pdf},
  note         = {Accessed: 2026-07-20}
}

@misc{CNECES2018Resolucao7,
  author       = {{Conselho Nacional de Educação} and {Câmara de Educação Superior}},
  title        = {Resolução CNE/CES nº 7, de 18 de dezembro de 2018: Estabelece as Diretrizes para a Extensão na Educação Superior Brasileira e regimenta o disposto na Meta 12.7 da Lei nº 13.005/2014, que aprova o Plano Nacional de Educação (PNE 2014--2024), e dá outras providências},
  year         = {2018},
  month        = dec,
  institution  = {Ministério da Educação},
  address      = {Brasília, DF, Brazil},
  url          = {https://www.gov.br/mec/pt-br/cne/pdf/resolucoes-do-cne/ces/2018/rces007_18.pdf},
  note         = {Accessed: 2026-08-04}
}

\end{document}